\documentclass[10pt,twocolumn,letterpaper]{article}

\usepackage{cvpr}              

\usepackage{graphicx}
\usepackage{amsmath}
\usepackage{amssymb}
\usepackage{booktabs}
\usepackage{multirow}
\usepackage{xcolor}
\newcommand{\light}[1]{\textcolor{gray}{#1}}

\usepackage[pagebackref,breaklinks,colorlinks]{hyperref}

\usepackage[capitalize]{cleveref}
\crefname{section}{Sec.}{Secs.}
\Crefname{section}{Section}{Sections}
\Crefname{table}{Table}{Tables}
\crefname{table}{Tab.}{Tabs.}

\def\cvprPaperID{*****} 
\def\confName{CVPR}
\def\confYear{2022}

\begin{document}

\title{Optimizing Anchor-based Detectors for Autonomous Driving Scenes}


\author{Xianzhi Du$^*$\\
Google
\and
Wei-Chih Hung\\
Waymo
\and
Tsung-Yi Lin\thanks{Work done while at Google}\\
Google
}

\maketitle

\begin{abstract}
This paper summarizes model improvements and inference-time optimizations for the popular anchor-based detectors in the scenes of autonomous driving. Based on the high-performing RCNN-RS and RetinaNet-RS detection frameworks designed for common detection scenes, we study a set of framework improvements to adapt the detectors to better detect small objects in crowd scenes. Then, we propose a model scaling strategy by scaling input resolution and model size to achieve a better speed-accuracy trade-off curve. We evaluate our family of models on the real-time 2D detection track of the Waymo Open Dataset (WOD)~\cite{waymo_open_dataset}. Within the 70 ms/frame latency constraint on a V100 GPU, our largest Cascade RCNN-RS model achieves 76.9\% AP/L1 and 70.1\% AP/L2, attaining the new state-of-the-art on WOD real-time 2D detection. Our fastest RetinaNet-RS model achieves 6.3 ms/frame while maintaining a reasonable detection precision at 50.7\% AP/L1 and 42.9\% AP/L2.

\end{abstract}

\section{Introduction}
\label{sec:intro}
Object bounding box detection in autonomous driving scenes is one of the most popular yet challenging tasks in computer vision. Unlike common object detection scenes that cover a wide range of detection scenarios, object scales and object categorizes, autonomous driving scenes typically focus on street driving views where objects of interest are smaller in size and are from less categories. Typical objects of interest for autonomous driving scenes are cars, pedestrians, cyclists, motorists, street signs, etc.

\begin{figure}
    \includegraphics[width=1.0\columnwidth]{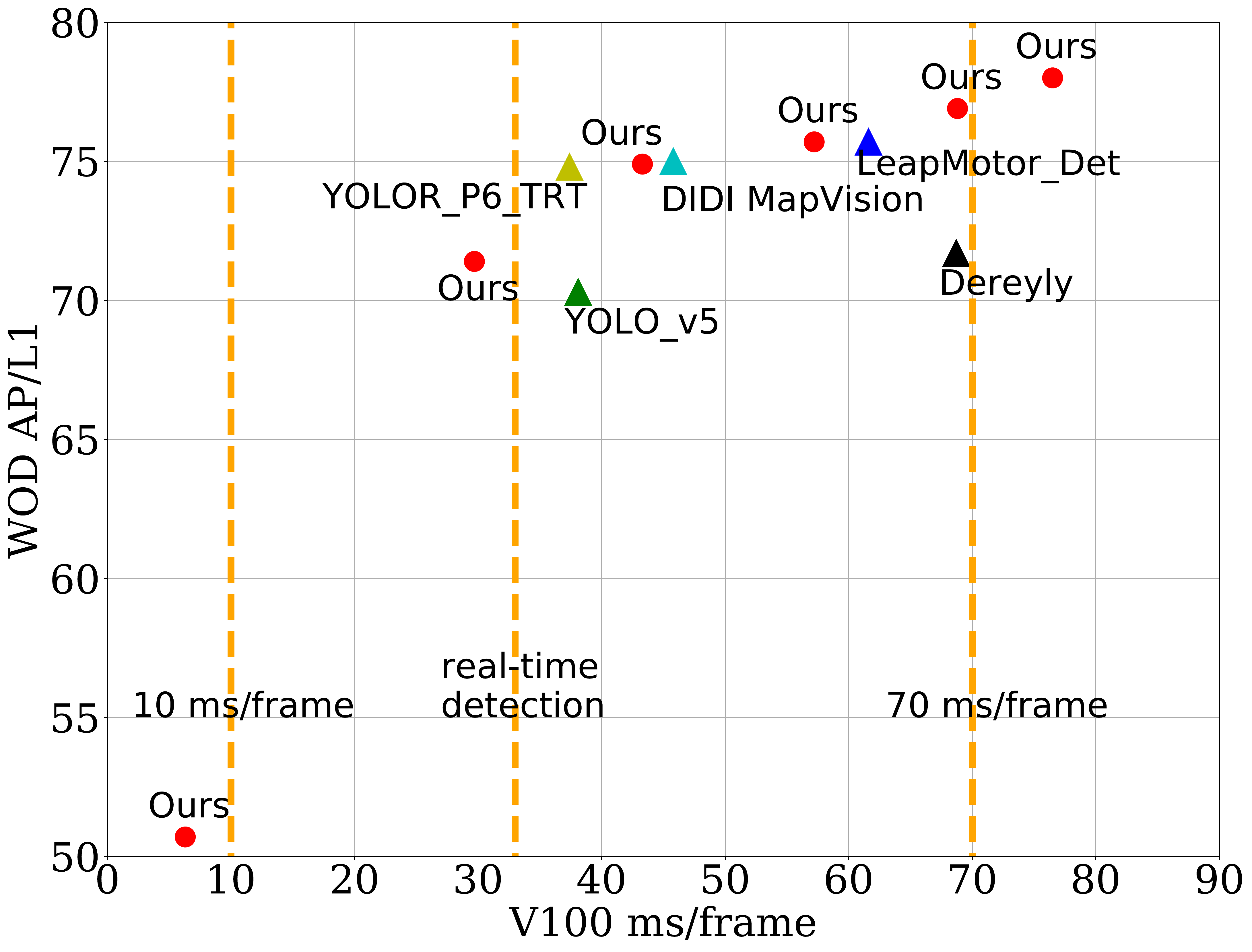}
    \caption{Result comparison on the WOD real-time 2D detection leaderboard. Given the 70 ms/frame latency constraint on a V100 GPU, our best model achieves the new state-of-the-art performance at 76.9\% AP/L1. More details can be found in Section~\ref{sec:exp}.}
    \label{fig:pareto_curve}
\vspace*{-5mm}
\end{figure}

The COCO~\cite{coco} benchmark has been the de facto benchmark to evaluate performance of object detectors since 2015. As a result, most of the popular object detectors~\cite{ssd, rcnn, fast_rcnn, fasterrcnn, mrcnn, cai2018cascade, retinanet, fpn, nasfpn, Du2020SpineNetLS, Du2020EfficientSB, efficientdet, du2021simple, bochkovskiy2020yolov4, liu2021swin} are tailored for COCO detection in model design, training recipe, post-processing methods, inference-time optimization, and model scaling strategy. In this work, we aim to optimize common object detectors for autonomous driving. We adopt the RCNN-RS and RetinaNet-RS~\cite{du2021simple} frameworks as our strong baselines and carefully study the effectiveness of tuning common COCO detection settings for autonomous driving scenes in model improvements and inference-time optimizations. Next, we discover a better strategy to scale models in input resolution and backbone scales and propose a family of models that form a better speed-accuracy trade-off curve. 

Our family of models are evaluated on the real-time 2D detection track of the Waymo Open Dataset~\cite{waymo_open_dataset} (WOD). We adopt the RCNN-RS models as our main detectors to push for best accuracy and adopt the RetinaNet-RS models to push for fastest speed. Under the 70 frame/ms latency constraint on a V100 GPU, our Cascade RCNN-RS model achieves 76.9\% AP/L1 and runs at 68.8 frame/ms, achieving the new state-of-the-art on the real-time 2D detection leaderboard~\cite{wod_2d_leaderboard}. To further push for highest accuracy and fastest speed, our largest Cascade RCNN-RS model achieves 78.9\% AP/L1 and runs at 103.9 ms/frame and our smallest RetinaNet-RS achieves 6.3 ms/frames while maintaining a reasonable AP/L1 at 50.7\%, respectively. 


\setlength{\tabcolsep}{4pt}
\begin{table}[h]
\begin{center}{
\begin{tabular}{l | l | l}
\toprule
 Model & Lat (ms/frame) & AP\slash L1\\
 \midrule
 Faster RCNN-RS &  91.8 & 71.5\\
 + 3 cascaded heads & 122.5 & 72.8 \textcolor{blue}{(+1.3)}\\
 + L2-L6 features & 135.1 & 74.0 \textcolor{blue}{(+1.2)}\\
 + Lightweight heads & 100.8 \textcolor{blue}{(-25\%)} & 73.7 \\
 \midrule
 + 512 proposals  & 91.7   \textcolor{blue}{(-9\%)} & 73.7\\
 + 0.7 NMS threshold  & 91.7 & 74.9 \textcolor{blue}{(+1.2)}\\
 + TensorRT\&float16 & 43.3 \textcolor{blue}{(-53\%)}& 74.9\\

\bottomrule
\end{tabular}}
\end{center}
\vspace{-3mm}
\caption{Ablation studies of the RCNN-RS model improvements and inference-time optimizations on WOD 2D detection. By applying all the changes to the Faster RCNN-RS baseline, the final model achieves \textcolor{blue}{+3.4\%} AP/L1 while being \textcolor{blue}{53\%} faster.}
\label{tab:arc_changes}
\end{table}

\section{Methodology}
\subsection{RCNN-RS improvements}~\label{sec:rcnn_improve}
\vspace{-7mm}
\subsubsection{Architectural improvements}
We adopt the strong RCNN-RS~\cite{du2021simple} as our main detection framework. RCNN-RS provides improved object detection performance in common detection scenes by adopting modern training techniques and architectural improvements. The modern techniques include scaling jittering augmentation, stochastic depth regularization~\cite{dropconnect}, a longer training schedule and SiLU activation~\cite{silu, swish}. To further optimize the detection framework for autonomous driving scenes, we make the following changes.

\paragraph{Cascaded heads:} The Cascade R-CNN framework~\cite{cai2018cascade} shows consistent accuracy improvements over the Faster R-CNN~\cite{fasterrcnn} baseline. In this work, we adopt the Cascade RCNN-RS (CRCNN-RS) as our main detection framework. Unlike COCO detection where best accuracy is achieved with two cascaded heads with higher IoUs~\cite{du2021simple}, here we adopt three cascaded detection heads with increasing foreground IoU thresholds $\{0.5, 0.6, 0.7\}$.

\paragraph{Lightweight heads:} We remove all convolutional layers in the detection heads and the RPN head but only keep the final fully connected layer for bounding box regression and classification. The lightweight head design significantly boosts model speed while achieving similar accuracy as the original head design which consists of 4 convolutional layers and one fully connected layer.

\paragraph{L2-L6 feature pyramid:} Detecting objects on a multi-scale feature pyramid is crucial to achieve good performance~\cite{fpn}. A common design choice for COCO detection is to construct a L3-L7 feature pyramid~\cite{retinanet,nasfpn,Du2020SpineNetLS,Du2020EfficientSB,Tan2020EfficientDetSA}. To better adapt the detector to localize and recognize objects in smaller scales, we add L2 features to the feature pyramid and remove L7 features.

\subsubsection{Inference-time optimizations}
\paragraph{Inference-time framework designs:} We feed 512 detection proposals instead of 1000 to the second-stage of the CRCNN-RS detector for inference and training. NMS threshold is increased from 0.5 to 0.7 for ROI generation and final detection generation.

\paragraph{Benchmarking improvements:} We further adopt the NVIDIA TensorRT optimization with \texttt{float16} precision to optimize model inference speed.

\subsection{RetinaNet-RS improvements}
RetinaNet(-RS)~\cite{retinanet, du2021simple}, a popular anchor-based one-stage object detector, shows competitive detection performance in the low-latency regime for common object detection. In this work, we adopt the RetinaNet-RS framework for our fastest models and apply the changes introduced in Section~\ref{sec:rcnn_improve} except the ones specific to RCNN-RS. The changes include L2-L6 features, a larger NMS threshold, TensorRT and \texttt{float16} inference precision.

\subsection{Model scaling on WOD}~\label{sec:scaling}
We explore model scaling on the WOD from scaling input resolution and scaling backbone size. A better speed-accuracy trade-off curve is formed by selecting best-performing models within a wide range of computational cost. To scale input resolution, we gradually increase the height of the input image from 384 to 1536 and the width from 640 to 2688. To scale backbone, we adopt architectures at 5 different scales: ResNet-RS-18$\times$0.25\footnote{$\times$0.25 denotes the model's channel dimension is uniformly scaled to 0.25 of the original size.}~\cite{bello2021revisiting} (RN18$\times$0.25) that contains 5.3M parameters; SpineNet-49$\times$0.25 (SN49$\times$0.25) that contains 5.6M parameters; SpineNet-49 (SN49) that contains 30.3M parameters; SpineNet-96 (SN96) that contains 37.6M parameters; SpineNet-143 (SN143) that contains 49.7M parameters. Table~\ref{tab:waymo} presents our models of all scales. 


\setlength{\tabcolsep}{4pt}
\begin{table*}[h]
\begin{center}{
\begin{tabular}{l | c | c c | c  c c}
\toprule
 Backbone & Input res &  Params (M) & FLOPs (B) & Lat (ms/frame) & AP\slash L1 & AP\slash L2\\
 \midrule
 RN18$\times$0.25 & 640$\times$1152 & 5.3& \light{9.3} & 13.7  & 59.6 & 51.3\\
 RN18$\times$0.25 & 768$\times$1408 & 5.3 & \light{11.4} & 16.6  & 62.1 & 53.9\\
 \midrule
 SN49$\times$0.25 & 640$\times$1152 & 5.6 & \light{10.4} & 19.6  & 63.9 & 55.5\\
 SN49$\times$0.25 & 768$\times$1408 & 5.6 & \light{13.0} & 22.5  & 66.1 & 58.1\\
 SN49$\times$0.25 & 896$\times$1664 & 5.6 & \light{16.0} & 26.2  & 67.9 & 60.0\\
 SN49$\times$0.25 & 1024$\times$1920 & 5.6 & \light{19.5} & 30.7  & 69.2 & 61.5\\
 \midrule
 \light{SN49} & \light{384$\times$640} & \light{30.3} & - & \light{18.8}   & \light{62.8} & \light{54.0}\\
 SN49 & 512$\times$896 & 30.3 & \light{57} & 22.0  & 68.3 & 59.9\\
 SN49 & 640$\times$1152 & 30.3 & \light{79} & 29.7  & 71.4 & 63.4\\
 SN49 & 768$\times$1408 & 30.3 & \light{107} & 34.9  & 73.4 & 65.7 \\
 SN49 & 896$\times$1664 & 30.3 & \light{141}& 43.3  & 74.9 & 67.5\\
 SN49 & 1024$\times$1920 & 30.3 & \light{180}& 57.2 & 75.7 & 68.6\\
 SN49 & 1280$\times$2176 & 30.3 & \light{244}& 68.8 & 76.9 &70.1\\
 \light{SN49} & \light{1408$\times$2432} & \light{30.3} & \light{315}& \light{93.8}  &\light{77.3} & \light{70.4} \\
 \light{SN49} & \light{1536$\times$2688} & \light{30.3} & \light{372}& \light{97.1}  &\light{77.7} & \light{70.9}\\
 \midrule
 SN96 & 1280$\times$2176 & 37.6 & \light{381} & 76.5  & 78.0 & 71.2\\
 \midrule
 SN143 & 1280$\times$2176 & 49.7 & \light{587} & 103.9  &78.9 & 72.3\\

\bottomrule
\end{tabular}
}
\end{center}
\vspace{-3mm}
\caption{\textbf{CRCNN-RS model performance on WOD.} We study model scaling on WOD by scaling up input resolution and backbone size. All models adopt the CRCNN-RS framework.}
\label{tab:waymo}
\end{table*}

\setlength{\tabcolsep}{4pt}
\begin{table}[h]
\begin{center}{
\begin{tabular}{l |c c c}
\toprule
 Input res & Lat (ms/frame) & AP\slash L1 & AP\slash L2\\
 \midrule
  512$\times$896 &  6.3  & 50.7 & 42.9 \\
 640$\times$1152 &  11.6  & 54.2 & 46.5 \\
 768$\times$1408 &  13.6 & 55.5 & 48.0\\

\bottomrule
\end{tabular}
}
\end{center}
\vspace{-3mm}
\caption{\textbf{RetinaNet-RS performance on WOD.} All models adopt a RN18$\times$0.25 backbone.}
\label{tab:retina}
\end{table}

\section{Experimental Results}~\label{sec:exp}
\vspace{-5mm}
\subsection{WOD training and eval settings}
We conduct experiments on the real-time 2D detection track of the popular Waymo Open Dataset~\cite{waymo_open_dataset}. WOD is a large-scale dataset for autonomous driving that consists of 798~training sequences and 202~validation sequences. Each sequence spans 20~seconds and is densely labeled at 10~frames per second with camera object detection and tracks.

We train all models on the WOD \texttt{train} split with synchronized batch normalization, SGD with a 0.9 momentum rate and a batch size of 256 for 20000 steps on TPUv3 devices~\cite{jouppi2017tpu}. We apply a cosine learning rate schedule with an initial learning rate 0.32. A linear learning rate warm-up is applied for the first 1000 steps. To obtain competitive results, we pretrain our models on the COCO~\cite{coco} dataset by following the training practices from~\cite{du2021simple}. Our main results for 2D bounding box detection are reported on the WOD \texttt{test} split.

\subsection{Improvements from architecture and inference-time optimization}
In this section, we show the impact of the RCNN-RS model architectural changes and inference-time model optimizations for autonomous driving scenes. The detailed ablation studies are shown in Table~\ref{tab:arc_changes}. For the architectural changes, starting from the Faster RCNN-RS model, adding two more cascaded heads increases AP/L1 by \textcolor{blue}{+1.3\%}. Introducing L2 features to the multiscale feature pyramid increases AP/L1 by another \textcolor{blue}{+1.2\%}. Removing $3\times3$ convolutional layers in the RPN head and the detection heads speeds up the model by \textcolor{blue}{25\%} while achieving similar accuracy. For the inference-time optimizations, by reducing the number of proposals for the second stage from 1000 to 512 and increasing the NMS threshold from 0.5 to 0.7, we further improve AP/L1 by \textcolor{blue}{+1.2\%} while being \textcolor{blue}{9\%} faster. Finally, optimizing model with TensorRT and changing inference model precision from \texttt{float32} to \texttt{float16} significantly reduces inference latency by (\textcolor{blue}{53\%}).

\subsection{Scaling input resolution \vs backbone size}
\begin{figure}
    \includegraphics[width=1.0\columnwidth]{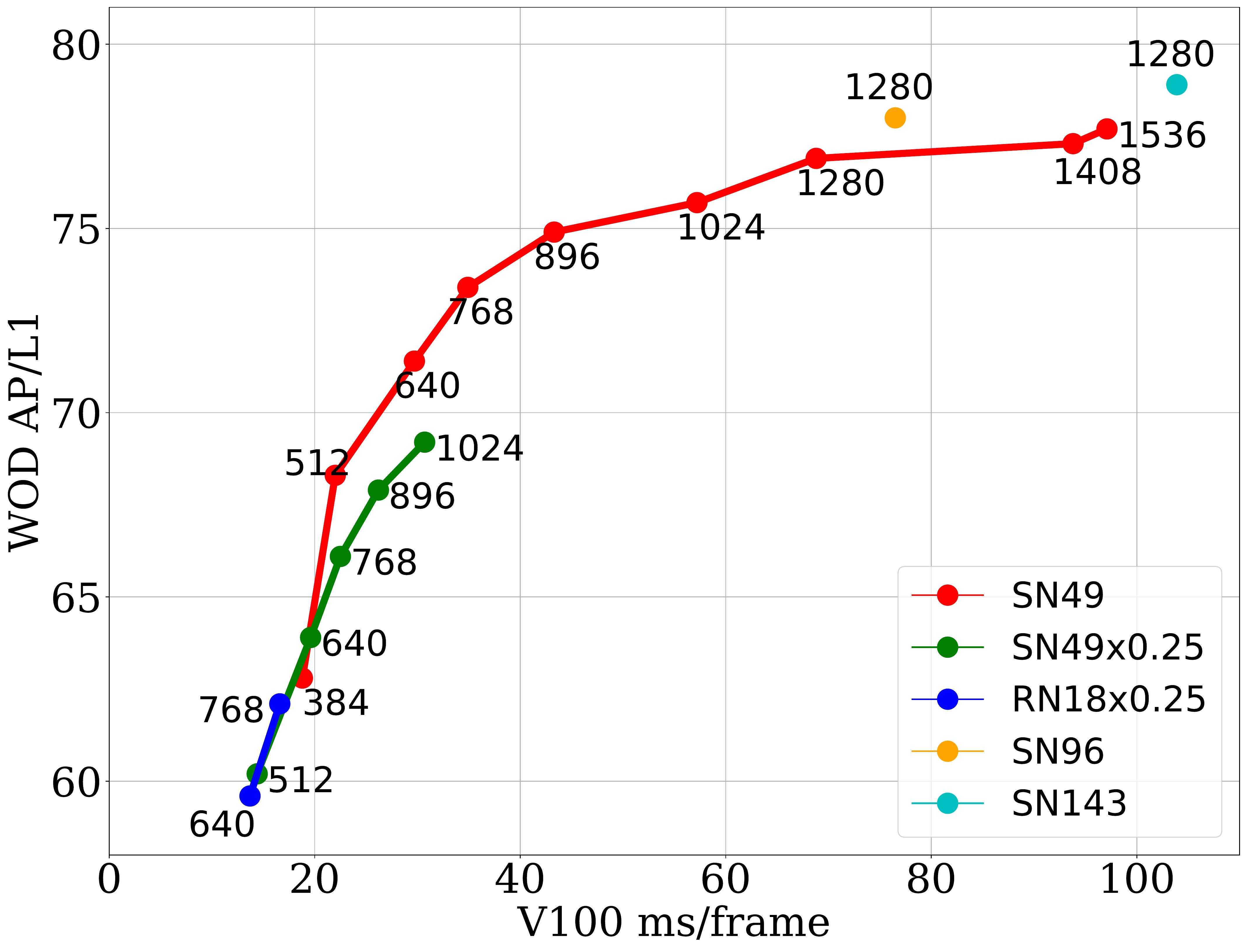}
    \caption{\textbf{Model scaling on WOD 2D detection.} We compare CRCNN-RS models adopting SN49, SN96, SN143, SN49$\times$0.25 and RN18$\times$0.25 backbones at various input resolutions. Numbers in this figure represent height of the input image.}
    \label{fig:scaling_curve}
\vspace*{0mm}
\end{figure}

We explore the effectiveness of scaling input resolution vs. scaling backbone size by a grid search over the scales described in Section~\ref{sec:scaling} for CRCNN-RS models. The results are presented in Table~\ref{tab:waymo} and Fig.~\ref{fig:scaling_curve}. In Fig.~\ref{fig:scaling_curve}, we show that within a reasonable input resolution range from height 512 to 1280, scaling input resolution while using a larger backbone is more effective than adopting a smaller backbone with larger input resolutions. To further push for a higher accuracy or a faster speed, scaling backbone size while keeping input resolution becomes a more effective strategy.

\subsection{RCNN \vs RetinaNet in the low-latency regime}
\begin{figure}
    \includegraphics[width=1.0\columnwidth]{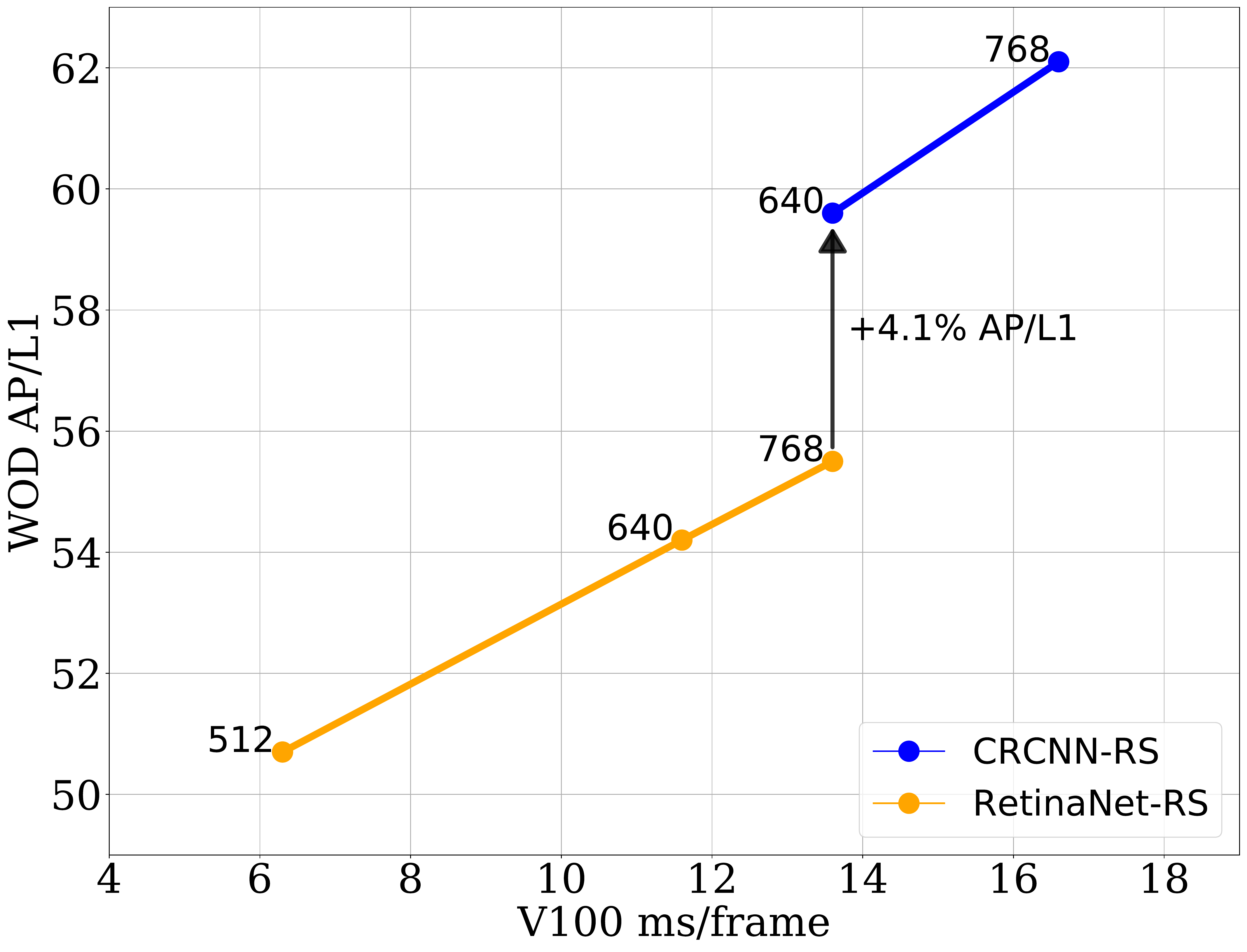}
    \caption{\textbf{RetinaNet-RS \vs CRCNN-RS in the low-latency regime.} All models adopt a RN18$\times$0.25 backbone. Numbers in this figure represent height of the input image.}
    \label{fig:retina_curve}
\vspace*{0mm}
\end{figure}

In this section, we evaluate the performance comparison between RCNN-RS and RetinaNet-RS on WOD. We apply the same training and benchmarking practices. The results are shown in Table~\ref{tab:retina}. We compare the RetinaNet-RS models to the CRCNN-RS models adopting a RN18$\times$0.25 backbone. As shown in Fig.~\ref{fig:retina_curve}, the RetinaNet-RS model underperforms CRCNN-RS by 4.1\% AP/L1 while running at a same speed. On the other hand, benefiting from the one-stage framework design, RetinaNet-RS achieves the fastest speed at 6.3 ms/frame. 

\vspace{10mm}

\vspace{-7mm}
\subsection{WOD Real-time 2D detection results}
We present our best performing models and show the performance comparisons to the top-5 entries on the real-time 2D detection leaderboard~\cite{wod_2d_leaderboard} in Table~\ref{tab:sota_models} and Fig.~\ref{fig:pareto_curve}. In particular, within the 70 ms/frame constraint, our CRCNN-SN49 model at 1280$\times$2176 input resolution achieves 76.9\% AP/L1 and 70.1\% AP/L2 and runs at 68.8 ms/frame, outperforming previous best models on the leaderboard. Our CRCNN-SN49 model at 640$\times$1152 input resolution achieves 71.4\% AP/L1 and 63.4 AP/L2 and runs at 29.7 ms/frame, achieving real-time object detection whiling attaining competitive detection precision. In the low-latency regime, our smallest RetinaNet-RS adopting a RN18$\times$0.25 backbone at 512$\times$896 resolution achieves 6.3 ms/frame while maintaining reasonable detection precision at 50.7\% AP/L1 and 42.9\% AP/L2. 

\setlength{\tabcolsep}{4pt}
\begin{table}[h]
\begin{center}{
\begin{tabular}{l  | c | c c}
\toprule
 Model & Lat. (ms) & AP\slash L1 & AP\slash L2\\
 \midrule
 CRCNN-RS-SN49 @640 &  29.7  & 71.4 & 63.4\\
 CRCNN-RS-SN49 @896 &  43.3 & 74.9 & 67.5\\
 CRCNN-RS-SN49 @1024 &  57.2 & 75.7 & 68.6\\
 CRCNN-RS-SN49 @1280 & 68.8 & 76.9 & 70.1\\
 CRCNN-RS-SN96 @1280 & 76.5 & 78.0 & 71.2\\
 \midrule
 \midrule
 LeapMotor\_Det~\cite{LeapMotor_Det} &  61.6 & 75.7 & 70.4 \\
 DIDI MapVision~\cite{DIDI_MapVision} &  45.8  & 75.0 & 69.7 \\
 YOLOR\_P6\_TRT~\cite{YOLOR_P6_TRT} &  37.4 & 74.8 & 69.6 \\
 Dereyly\_self\_ensemble~\cite{dereyly_self_ensemble} & 68.7 & 71.7 & 65.7 \\
 YOLO\_v5~\cite{YOLO_v5} & 38.1 & 70.3 & 64.1 \\

\bottomrule
\end{tabular}
}
\end{center}
\vspace{-3mm}
\caption{Result comparisons of our models against the top-5 models on the WOD real-time 2D detection leaderboard. We omit results using model ensemble or multiscale test.}
\label{tab:sota_models}

\end{table}

\section{Conclusion}
In this work, we improve the strong two-stage RCNN-RS detector for autonomous driving scenarios from architectural changes and inference-time optimizations. We study the impact of scaling input resolution and model size on the task of WOD real-time 2D detection and propose a family of models for a wide range of latency. We hope this study will help the community to better design detectors for autonomous driving and the optimizations can transfer to more detection frameworks and detection scenarios.

\vspace{3mm}
\par\noindent\textbf{Acknowledgments:} We would like to acknowledge Henrik Kretzschmar, Drago Anguelov and the Waymo research team for the support. Barret Zoph, Jianwei Xie, Zongwei Zhou, Nimit Nigania for the helpful discussions.

{\small
\bibliographystyle{ieee_fullname}
\bibliography{egbib}
}

\end{document}